\documentclass{article}

\usepackage[final]{DLI_2023}

\usepackage[utf8]{inputenc} % allow utf-8 input
\usepackage[T1]{fontenc}    % use 8-bit T1 fonts
\usepackage{hyperref}       % hyperlinks
\usepackage{url}            % simple URL typesetting
\usepackage{booktabs}       % professional-quality tables
\usepackage{amsfonts}       % blackboard math symbols
\usepackage{nicefrac}       % compact symbols for 1/2, etc.
\usepackage{microtype}      % microtypography
\usepackage{xcolor}         % colors
\usepackage{graphicx}
\usepackage{graphicx}
\usepackage{subfig}

\title{A Video-based Detector for Suspicious Activity in Examination with OpenPose}

\author{%
  Reuben Moyo\\
  University of Dar es Salaam, Tanzania\\
  \texttt{reubencmoyo@gmail.com} \\
  \And
  Stanley Ndebvu \\
  University of Botswana, Botswana \\
  \texttt{stanndebvu@yahoo.com} \\
  \AND
  Michael Zimba \\
  Malawi University of Science \& Technology, Malawi \\
  \texttt{mgmzimba@gmail.com} \\
   \And
  Jimmy Mbelwa \\
  University of Dar es Salaam, Tanzania \\
  \texttt{jimmymbelwa@gmail.com} \\
}

\begin{document}

\maketitle

\begin{abstract}
  Examinations are a crucial part of the learning process, and academic institutions invest significant resources into maintaining their integrity by preventing cheating from students or facilitators. However, cheating has become rampant in examination setups, compromising their integrity. The traditional method of relying on invigilators to monitor every student is impractical and ineffective. To address this issue, there is a need to continuously record exam sessions to monitor students for suspicious activities. However, these recordings are often too lengthy for invigilators to analyze effectively, and fatigue may cause them to miss significant details. To widen the coverage, invigilators could use fixed overhead or wearable cameras. This paper introduces a framework that uses automation to analyze videos and detect suspicious activities during examinations efficiently and effectively. We utilized the OpenPose framework and Convolutional Neural Network (CNN) to identify students exchanging objects during exams. This detection system is vital in preventing cheating and promoting academic integrity, fairness, and quality education for institutions.
\end{abstract}

\section{Introduction}
The use of advanced technology in education has had a significant impact on the growth of the field. Technology is utilized in teaching, learning, and exam administration [1], among other areas. Exams are a crucial part of a student's learning, and academic institutions put in a lot of effort and resources to ensure the integrity of the exam. Despite the presence of a proctor, students may still cheat cautiously to avoid harsh consequences[2]-[4]. We can record and analyse suspicious behaviour during exam sessions using security cameras and OpenPose, a real-time multi-person 2D pose estimation library [5]. These videos can contain information that even invigilators may miss. However, going through each video manually is time-consuming and arduous. Automating the analysis and evaluation of the videos and highlighting any suspicious activities would be helpful. Most modern lecture halls have cameras installed in strategic locations for security purposes. Similarly, these cameras can record and analyze any unusual or suspicious activity during exams, along with wearable cameras if necessary. This paper discusses a framework that automatically analyses pre-recorded exam videos and detects suspicious activities like exchanging objects among students during an examination.

Several techniques that students use to cheat during exams are highlighted by Anderman [6], Cavalcanti et al. [7], and Abdaoui [8]. These include talking to each other, shaking hands, and passing objects and materials. Atoum [1] and D'Souza [9] suggest a framework teachers can use to detect cheating by their students. Our proposed solution uses advanced computer vision, deep learning, and video content analysis techniques and models to analyze pre-recorded videos and detect cheating during exams. The posture and orientation of the human body [10], [11] play a crucial role in identifying students who are talking to each other, shaking hands, and passing objects and materials. We can also detect an individual's sitting position during an exam to identify suspicious activity. Cao et al. [12] discuss the Top-down and Bottom-up methods for Multi-person pose estimation to locate human joints and determine whose joint it is. Chen [10] recommends using computer vision-based sitting posture estimation, which is more cost-effective and does not require additional hardware compared to sensor-based methods.

\subsection{Objectives}
This study intended to analyse pre-recorded videos of an exam session in a controlled exam hall and detect suspicious activities like students exchanging objects in an exam to assist a physical invigilator in continuously monitoring the students.

\section{Methodology}
A class was formed by randomly selecting sixteen (16) students who participated in an experiment by recording videos. We arranged the students in four rows and columns to create a natural exam environment. The various activities of the participants are presented in Table 1.

Ten (10) trials of ten-minute videos were conducted, with the first six runs consisting of four specific actions in each run. Two runs were used as a control where all students adhered to strict exam rules. In the last four runs, participants performed random activities. As a result, we obtained a smaller dataset of 10,000 images, which we divided into 50\% for the training set, 25\% for the validation set, and 25\% for the testing set. Frames labelled with shaking hands depict a scenario where two students pretend to greet each other while they exchange something in their palms. The structures labelled passing\_object represent a scenario where students exchange objects.

 \begin{table}
      \centering
      \begin{tabular}{l r r c}
        \toprule
        \textbf{Action} & \textbf{Suspected Act} \\
        \midrule
        Shake\_Hands & Passing of a piece of paper \\
        Exchange\_Object & Passing exam scripts  \\
        Use\_Phone & Use of reference material  \\
        Throw\_Object & Form of exchanging objects  \\
        \bottomrule
      \end{tabular}
      \caption{Activities conducted during the experiment by the students. We derived metrics for evaluation from these activities.}
    \end{table}

\subsection{Design and Evaluation}
\begin{figure}
      \centering
      \includegraphics[height=1.5in, width=2.0in]{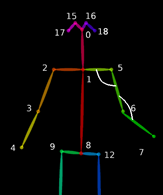}
       \caption{An extract of 25 body key points for OpenPose focusing on key points of interest in the upper part of the body.}
      \label{fig:pic1}%
    \end{figure}

Our design uses the OpenPose framework to extract feature maps with 25 body key points where each key point consists of x and y geometric coordinates and a confidence value. We load the OpenPose pre-trained model and run it on the pre-recorded videos to extract the key points of significance: Wrist, Elbow, Shoulder, and Neck. We further categorised the key points into Left\_Wrist (4), Left\_Elbow (3), Left\_Shoulder (2), Neck (1), Right\_Wrist (7), Right\_Elbow (6), and Right\_Shoulder (5); see Figure~\ref{fig:pic1}. 

We used the 2D coordinates of key points to create a geometric angle between two vectors. This allowed us to implement a geometric evaluation to determine if a hand is stretched sideways or not. In their papers, "Human Activity Recognition with OpenPose and Long Short-Term Memory on Real-Time Images" and "An OpenPose Based Method to Detect Texting While Walking", Sawant et al. [46] and Hanaizumi et al. [13] present a similar approach to recognize human activity. They use angles formulated from key points to detect if a person is walking while using their phone. Our pipeline is based on their work, and we used it to recognize if a student is stretching a hand sideways to receive an object from another student. Our pipeline focuses on extracting a skeletal framework of body key points using the OpenPose library and vectorizing the coordinates of the key points we are interested in. We draw a skeletal framework over each detected student, and the evaluation is done per student.  
For our experiment, we chose common activities that students engage in, such as shaking hands, using phones, throwing objects, and exchanging objects. We selected these actions based on feasibility and Adrian's recommendations within a limited time frame [14]. We used pre-recorded videos and applied an OpenPose framework for pose estimation detection. We recorded the x and y geometric coordinates and confidence value of each key point and then reconstructed a skeletal estimation of a person by joining the key points. We identified and tracked the person extending their hand, focusing on key points in the wrist, elbow, shoulder, and neck to form forearm and upper arm vectors. To account for variations in body length measurements and distance from the camera, we used linear interpolation as described in [15]. The coordinates of human body key points may fluctuate due to recognition uncertainty, so we modified canonical correlation analysis to smooth out slight changes. 

We produced a projection vector from two absolute coordinates \(A \in R^{m}\) and \(B \in R^{n}\) where a distance matrix is constructed \(D \in R^{mn}\) where \(D_{i,j}\) is the distance between \(A_i\) and \(B_j\). After constructing projection vectors from the coordinates, we applied a Dot Product effect on the vectors to find a relative angle between vectors which is 
given by \[P\cdot Q = \sum_{k=1}^{n}{p_iq_i}\] where \(P=(p_1,p_2)\) and \(Q=(q_1,q_2)\) are 2D vectors. 

Upon constructing the forearm vector, and the upper arm vector, we find the relative angle between the two vectors by the given inverse cosine:
\[\theta=\arccos(\frac{P\cdot Q}{\left| P \right|\left| Q \right|})\] where \(0 \leq \theta\leq 180\) as the maximum angle an arm can extend to is 180 degrees. When a student raises their arm to the side, the angle between their upper arm and neck increases, as well as the angle between their upper arm and forearm. This happens because of the correlation that exists among these limbs of the body. We then analyze the angles and determine a threshold angle based on the data. We can identify any outlier angle values by calculating each person's Standard Deviation (SD). These outlier values indicate the possibility of... falling within one SD. 
Therefore, to deduce an extended arm, we applied the following: \[f(x,y)= \left\{ \begin{array}{cl}
x \geq T \leq 180\\ y \geq 90 \end{array} \right.\]
                        
where T is the Threshold, x is the angle between the forearm vector and upper arm vector, and y is the angle between the upper arm vector and the neck vector. 
When a student extends a hand in shaking-hands, passing-object, and throwing-object, the angle between the neck vector and the upper arm vector increases, and we intend to use such geometric heuristics to determine the stretching of the hand.

\section{Results and Discussion}
\begin{figure}%
    \centering
    \subfloat[\centering label A]{{\includegraphics[width=5cm]{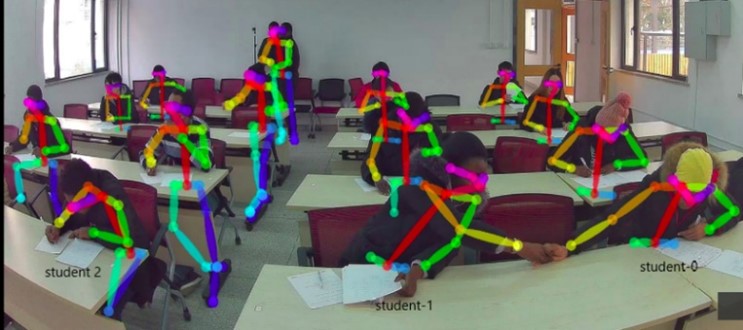} }}%
    \qquad
    \subfloat[\centering label B]{{\includegraphics[width=5cm]{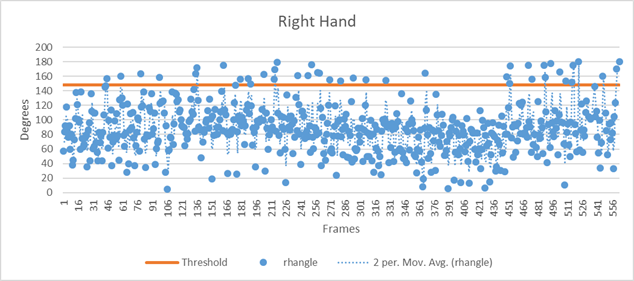} }}%
    \caption{Student-0 and student-1 exchange materials in an uncontrolled state experiment at time z}%
    \label{fig:studentexchange}%
\end{figure}

We utilized the multi-person OpenPose framework to thoroughly analyze recorded videos, pinpointing the key points of each of the 13 students in each frame. Our analysis included extracting key points for the left wrist, left elbow, left shoulder, neck, right wrist, right elbow, and right shoulder, with each key point providing both geometric coordinates (x and y) and a confidence value. With vectorization, we could then accurately calculate the angles needed for our analysis illustrated in Figure ~\ref{fig:studentexchange} (a).

\textbf{Scenario 1: Overall pattern in a controlled state:} During the first test scenario, the students had to follow strict rules that prohibited stretching their hands, throwing or exchanging objects. Figure ~\ref{fig:studentexchange} (b) shows a scatter graph representing the right-hand and right-shoulder angles of 13 students. It was observed that when a person extended their hand to receive an object, the angle of the right-hand increased, as did the right shoulder angle. However, a person can receive or pass an object by lowering their arm, but still stretching the right-hand relative angle to above 148 degrees while keeping the right-shoulder angle as small as possible. To determine whether the stretched hand was suspicious, we also considered the duration of the hand-stretch. If the angle values in sequential frames were above the threshold T and lasted for several milliseconds, the stretching of the hand was deemed suspicious.

In Figure ~\ref{fig:studentexchange} (b) above, most right-hand angles are below 148 degrees which is the T. The distribution of the dots also indicates normal hand stretching activity. However, a few dots are above T, meaning some individuals stretched their hands. Even so, the distribution of the dots above T shows that the stretching was just an instant episode, as they do not cover several milliseconds and frames. The stretching was not sustained for several minutes. Nevertheless, on average, the class performed well as it adhered to the instructions.

\begin{figure}%
    \centering
    \subfloat[\centering Student-1 stretching hand while receiving a piece of paper.]{{\includegraphics[width=5cm]{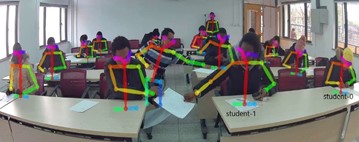} }}%
    \qquad
    \subfloat[\centering A scatter plot of the right-hand stretching pattern of student-1]{{\includegraphics[width=5cm]{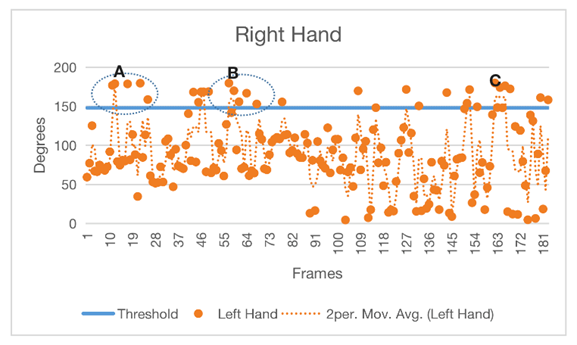} }}%
    \caption{A scatter plot of the right-hand stretching pattern of student-1}%
    \label{fig:uncontrolledstudent}%
\end{figure}

\textbf{Scenario 2: Student -1 Pattern in an Uncontrolled State:}. In the third test scenario, we analyzed the activities of a single student in an uncontrolled environment. The student was allowed to throw and exchange objects and stretch their hands. Figure ~\ref{fig:uncontrolledstudent} (a) and Figure ~\ref{fig:uncontrolledstudent} (b) depict the student's activities and a scatter graph of their right-hand movements, respectively. Figure 10 reveals that student-1 stretched their right-hand multiple times, with most episodes lasting several milliseconds. This is evident from the dots distributed above the T line. The canonical correlation analysis for the right-hand and right-shoulder plots indicates that student-1 kept their right-shoulder angle relatively small while receiving an object, which lowered their arm.

Figure ~\ref{fig:uncontrolledstudent} (b) highlights Regions A, B, and C as areas of possible suspicious activity due to the angles being above T and spanning several milliseconds. Unlike other angles above T, these regions only show a single episode. To confirm the suspicion, sustained angle values in multiple frames are needed. A significant decrease in both angles can occur when the arm is suddenly stretched or withdrawn. There is no set threshold for the right-shoulder angles, as they are not independent indicators of whether a student is extending their hand to receive or give an object. However, when the right-shoulder angle matches the arm angle, it can improve the accuracy of the conclusion.  

\subsection{Ethical Consideration}
In this study, ethical issues cannot be disregarded. One of the issues is privacy. The privacy of the students who participated in the experiment was significantly considered by restricting the scope for more domain-specific actionable solutions while keeping sight of the big-picture. All experiment subjects were provided with a consent form as stipulated by [16] and [17]

\section{Conclusion}

A framework that uses advanced technology to detect suspicious activities in exam rooms was created. The system can detect if someone is exchanging objects by analyzing the frequency and duration of hand movements. Although this approach may not always be accurate, it is still worth noting and reviewing by the invigilator. The OpenPose framework was utilized to generate key points for each student, which were then evaluated using geometric heuristics. However, this framework still requires significant improvements before it can be used on a larger scale. The development was based on a limited data set with pre-defined protocols, which may not be sufficient in a complex environment like an exam room where cheating methods constantly evolve. Thus, incorporating action recognition through Recurrent Neural Network (RNN) approaches with Machine Learning (ML) methods is crucial for better accuracy.

\section*{References}

\medskip

{
\small
[1]	Y. Atoum, L. Chen, A. X. Liu, S. D. H. Hsu, and X. Liu, {\it“Automated Online Exam Proctoring,”} IEEE Trans. Multimed., vol. 19, no. 7, pp. 1609–1624, 2017, doi: 10.1109/TMM.2017.2656064.

[2]	J. Sheard, Simon, M. Butler, K. Falkner, M. Morgan, and A. Weerasinghe, {\it“Strategies for Maintaining Academic Integrity in First-Year Computing Courses,”} in Proceedings of the 2017 ACM Conference on Innovation and Technology in Computer Science Education, 2017, pp. 244–249, doi: 10.1145/3059009.3059064.

[3]	K. Curran, G. Middleton, and C. Doherty, {\it“Cheating in Exams with Technology,”} Int. J. Cyber Ethics Educ., vol. 1, no. 2, pp. 54–62, 2011, doi: 10.4018/ijcee.2011040105.

[4]	S. Davis, Patrick, and Drinan, {\it Cheating In School: What We Know \& What We Can Do.} Wiley-Blackwell, 2009.

[5]	Z. Cao, G. Hidalgo Martinez, T. Simon, S.-E. Wei, and Y. A. Sheikh, {\it “OpenPose: Realtime Multi-Person 2D Pose Estimation using Part Affinity Fields,”} IEEE Trans. Pattern Anal. Mach. Intell., vol. XXX, no. XXX, pp. 1–1, 2019, doi: 10.1109/tpami.2019.2929257.

[6]	E. M. Anderman and T. B. Murdock, {\it “The Psychology of Academic Cheating,”} Psychol. Acad. Cheating, pp. 1–5, 2007, doi: 10.1016/B978-012372541-7/50002-4.

[7]	E. R. Cavalcanti, C. E. Pires, E. P. Cavalcanti, and V. F. Pires, {\it “Detection and evaluation of cheating on college exams using supervised classification,” } Informatics Educ., vol. 11, no. 2, pp. 169–190, 2012.

[8]	M. Abdaoui, {\it “Strategies for Avoiding Cheating and Preserving Academic Integrity in Tests,”} Alkhitab w el-Tawassol J., vol. 4, no. July 2018.

[9]	K. A. D’Souza and D. V Siegfeldt, {\it “A Conceptual Framework for Detecting Cheating in Online and Take-Home Exams,”} Decis. Sci. J. Innov. Educ., vol. 15, no. 4, pp. 370–391, 2017, doi: 10.1111/dsji.12140.

[10]	K. Chen, {\it “Sitting Posture Recognition Based on OpenPose,”} IOP Conf. Ser. Mater. Sci. Eng., vol. 677, no. 3, 2019, doi: 10.1088/1757-899X/677/3/032057.

[11]	J. C. T. Mallare et al., {\it “Sitting posture assessment using computer vision,”} HNICEM 2017 - 9th Int. Conf. Humanoid, Nanotechnology, Inf. Technol. Commun. Control. Environ. Manag., vol. 2018-Janua, pp. 1–5, 2017, doi: 10.1109/HNICEM.2017.8269473.

[12]	D. Osokin, {\it “Real-time 2D multi-person pose estimation on CPU: Lightweight OpenPose,”} ICPRAM 2019 - Proc. 8th Int. Conf. Pattern Recognit. Appl. Methods, pp. 744–748, 2019, doi: 10.5220/0007555407440748.

[13]	H. Hanaizumi and H. Misono, {\it “An OpenPose Based Method to Detect Texting while Walking,”} Proc. 7th IIAE Int. Conf. Intell. Syst. Image Process., pp. 130–134, 2019, doi: 10.12792/icisip2019.024.

[14]	S. Chen and R. Yang, {\it “Pose Trainer: Correcting Exercise Posture using Pose Estimation,”} no. March, 2018, doi: 10.13140/RG.2.2.29224.47367.

[15]	S. L. Colyer et al., {\it “Legal Implications of Using AI as an Exam Invigilator,”} SSRN Electron. J., vol. 75, no. 1, pp. 333–347, 2019, doi: 10.1111/hequ.12275.

[16]	S. Coghlan, T. Miller, and J. Paterson, {\it “Good proctor or ‘Big Brother’? AI Ethics and Online Exam Supervision Technologies,”} no. Ml, pp. 1–14, 2020, [Online]. Available: http://arxiv.org/abs/2011.07647.
}

%%%%%%%%%%%%%%%%%%%%%%%%%%%%%%%%%%%%%%%%%%%%%%%%%%%%%%%%%%%%

%%%%%%%%%%%%%%%%%%%%%%%%%%%%%%%%%%%%%%%%%%%%%%%%%%%%%%%%%%%%

\end{document}